\documentclass[journal]{IEEEtran}
\usepackage{cite}
\usepackage[pdftex]{graphicx}
\usepackage{amsmath}
\usepackage[ruled,vlined]{algorithm2e}
\usepackage{etoolbox}

\usepackage{siunitx}
\makeatletter
\patchcmd{\@algocf@start}
{-1.5em}
{0pt}
{}{}
\makeatother

\usepackage{array}
\usepackage{url}

\usepackage[caption=false,font=footnotesize]{subfig}
\usepackage{float}

\usepackage[nolist]{acronym}
\begin{acronym}[DEN-ng]
\acro{CAN}{Cognitive Autonomous Network}
\acro{EMA}{Environmental Modeling and Abstraction}

\acro{ACC}{clustering ACCuracy}
\acro{NMI}{Normalized Mutual Information}
\acro{ARI}{Adjusted Rand Index}

\acro{DEC}{Deep Embedded Clustering}
\acro{DEPICT}{DEeP embedded	regularIzed ClusTering}
\acro{VaDE}{Variational Deep Embedding}
\acro{ACAI}{Adversarially Constrained Autoencoder Interpolation}
\acro{ClusterGAN}{Latent Space Clustering in Generative Adversarial Networks}
\acro{IMSAT}{Information Maximizing	Self-Augmented Training}
\acro{RIM}{Regularized Information Maximization}
\acro{DAC}{Deep Adaptive image Clustering}
\acro{DCCS}{Deep image Clustering with Category-Style representation}
\acro{IIC}{Invariant Information Clustering}
\acro{ADC}{Associative Deep Clustering}
\acro{JULE}{Joint Unsupervised LEarning}

\acro{DANCE}{Decorrelating Adversarial Nets for Clustering-friendly Encoding}

\acro{GAN}{Generative Adversarial Net}
\acro{MSE}{Mean Squared Error}

\acro{DAN}{Decorrelating Adversarial Net}

\acro{QoE}{Quality of Experience}
\acro{KPI}{Key Performance Indicator}
\acro{VoIP}{Voice over IP}
\acro{FTP}{File Transfer Protocol}
\acro{HTTP}{HyperText Transfer Protocol}
\acro{CQI}{Channel Quality Indicator}
\acro{SNR}{Signal-to-Noise Ratio}
\acro{RSRP}{Reference Signals Received Power}
\acro{RRC}{Radio Resource Control}
\acro{RLF}{Radio Link Failure}

\acro{SAT}{Self-Augmented Training}

\end{acronym}

\usepackage{color, colortbl}

\def\BibTeX{{\rm B\kern-.05em{\sc i\kern-.025em b}\kern-.08em
		T\kern-.1667em\lower.7ex\hbox{E}\kern-.125emX}}

\newcommand{\kmeans}{\textit{k}-Means}

\definecolor{one}{RGB}{193, 213, 247}
\definecolor{two}{RGB}{196, 248, 209}

\newcommand{\hlone}[1]{\cellcolor{one}\textcolor{black}{#1}}
\newcommand{\hltwo}[1]{\cellcolor{two}\textcolor{black}{#1}}

\usepackage{lipsum}
\newcommand{\refhold}[1]{\textbf{!!REF!!}}

\begin{document}

\title{Decorrelating Adversarial Nets for Clustering \\ Mobile Network Data}

\author{
	\IEEEauthorblockN{
		M{\'a}rton Kaj{\'o}\IEEEauthorrefmark{1}\IEEEauthorrefmark{3},
		Janik Schnellbach\IEEEauthorrefmark{2}\IEEEauthorrefmark{3},
		Stephen S. Mwanje\IEEEauthorrefmark{3},
		Georg Carle\IEEEauthorrefmark{1}\\
	}
	\IEEEauthorblockA{
		\IEEEauthorrefmark{1}Technical University of Munich, Department of Informatics, Email: \{kajo, carle\}@net.in.tum.de\\
	}
	\IEEEauthorblockA{
		\IEEEauthorrefmark{2}Technical University of Munich, Email: janik.schnellbach@tum.de\\
	}
	\IEEEauthorblockA{
		\IEEEauthorrefmark{3}Nokia Bell Labs, Email: \{marton.kajo.ext, janik.schnellbach.ext, stephen.mwanje\}@nokia-bell-labs.com
	}
}


\maketitle

\begin{abstract}
	Deep learning will play a crucial role in enabling cognitive automation for the mobile networks of the future.
	Deep clustering, a subset of deep learning, could be a valuable tool for many network automation use-cases.
	Unfortunately, most state-of-the-art clustering algorithms target image datasets, which makes them hard to apply to mobile network data due to their highly tuned nature and related assumptions about the data.
	In this paper, we propose a new algorithm, \ac{DANCE}, intended to be a reliable deep clustering method which also performs well when applied to network automation use-cases. 
	\ac{DANCE} uses a reconstructive clustering approach, separating clustering-relevant from clustering-irrelevant features in a latent representation.
	This separation removes unnecessary information from the clustering, increasing consistency and peak performance.
	We comprehensively evaluate \ac{DANCE} and other select state-of-the-art deep clustering algorithms, and show that \ac{DANCE} outperforms these algorithms by a significant margin on a mobile network dataset.
\end{abstract}

\begin{IEEEkeywords}
	clustering, cognitive network automation, deep learning, unsupervised learning
\end{IEEEkeywords}

\acresetall{}

\section{Introduction}
	\label{sec:intro}
	
	\IEEEPARstart{D}{eep} learning algorithms are the basis for many of today's technological advancements, and will continue to enable future use-cases in all technological fields.
	Naturally, mobile network automation follows this trend, where future \acp{CAN} are expected to utilize deep learning in many network management tasks in order to make the network robust, reliable and adaptive \cite{can}.
	
	The majority of deep learning research targets supervised learning, such as classification, where the deep learning algorithms learn to output the ground truth that is explicitly defined in the training data in the form of labels or values, usually collected through crowdsourcing or data mining.
	However, these label generation processes are not available for mobile network automation, as network management tasks require expert knowledge, thus only a handful of people are capable of undertaking them.
	Requiring these experts to manually generate examples to be able to train deep learning algorithms in a supervised way is not feasible.
	This paper focuses on deep clustering, a form of unsupervised learning where the ground truth is not known, and the algorithm is not supported with the correct answers during training.
	In clustering, the task is to assign observations (data points) to groups (or clusters) based on their characteristics, so that similar observations end up in the same cluster.
	
	In the \ac{CAN} concept, an important element called \ac{EMA} uses clustering to define network states; discrete descriptions of the settings and performance of the network or parts thereof which are then used as the basis for decision making by the different \ac{CAN} components \cite{ema}.
	The correct designation of these \ac{EMA} network states is critical in this setting, as optimization, self-healing and load balancing decisions are mapped to the \ac{EMA} states directly.
	If an \ac{EMA} state incorporates dissimilar behavior, these control functions will not trigger at the correct place or in the correct time, making the network less reliable and less optimized.
	However, network states can not be defined explicitly, as a network's behavior changes depending on a myriad of internal and external factors, such as time, location, components used in the network, user behavior, etc.
	To this end, clustering is the only option to automatically define the \ac{EMA} network states.
	
	Apart from future management use-cases such as \ac{EMA}, clustering is already used in a variety of network and service management use-cases, where deep clustering could improve current functionality.
	Some examples of these are:
	\begin{itemize}
		\item Slice provisioning (instantiation) may utilize well-defined usage types through clustering, in order to select appropriate templates for the slices based on predicted requirements \cite{slice}.
		\item \ac{QoE} estimation tries to map explicitly measurable \acp{KPI}, such as network delay, jitter or throughput to user satisfaction levels \cite{qoe}. Here, clustering may be utilized to establish user archetypes.
		\item Network anomaly detection may utilize generative clustering to map normal usage patterns, and detect outliers which point to anomalous events in the network \cite{anomaly}.
	\end{itemize}

	\begin{figure}[t]
		\centering
		\includegraphics[width=\linewidth]{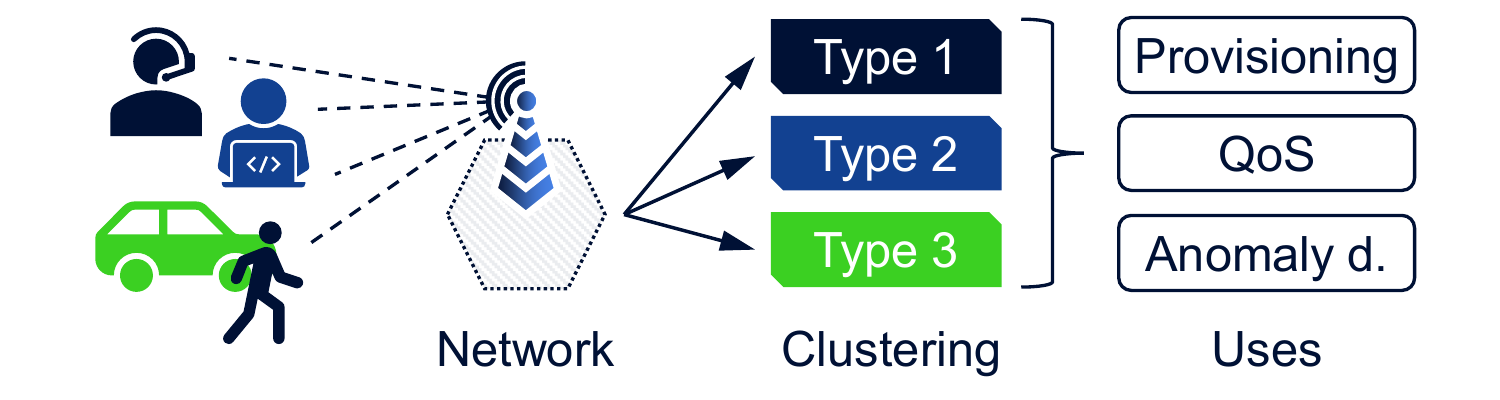}
		\caption{Network automation use-cases involving behavior clustering.}
		\label{fig:use_cases}
	\end{figure}
	
	These tasks all involve clustering of the \textit{behavior} of users, applications or services, by finding implicit patterns in the collected data (Fig. \ref{fig:use_cases}).
	Feature engineering and manual definition of clustering rules is quite hopeless, as the generally collected information that could be useful for these tasks does not contain the necessary information explicitly.
	Well-intended targeted data collection, such as deep packet inspection (snooping), is also increasingly difficult, as more and more of the communication is encrypted or governed by privacy laws (and rightly so).
	Thus, deep clustering algorithms that simultaneously extract implicit patterns and form groups using these patterns are the perfect match for these tasks.
	
	Deep clustering has seen a surge in attention recently, with huge performance improvements being published every few months.
	Some algorithms are now quite close to supervised classification performance, a feat that was unimaginable even a few years ago.
	However, as we show in this paper, applying these cutting-edge algorithms to mobile network data is not straight-forward.
	Most deep clustering algorithms are developed for image datasets, and are able to achieve great performance because of inherent assumptions and optimization that are specific to image data.
	These biases often don't translate well to mobile networks, where the performance of the algorithms degrades, in some cases significantly.
	We will elaborate on the challenges faced in applying these new algorithms in network automation and how they can be overcome.

	In this paper:
	\begin{itemize}	
		\item We discuss the design philosophy and examine the performance of state-of-the-art deep clustering algorithms (Sec. \ref{sec:sota}), highlighting their strengths and weaknesses.
		\item We propose our own deep clustering algorithm (Sec. \ref{sec:dance}), which aims to mitigate some of the weaknesses of state-of-the-art algorithms. 
		Furthermore, our algorithm is not biased to any application-domain, making it a preferable choice for clustering mobile network data.
		\item We evaluate our algorithm extensively by comparing its performance against state-of-the-art deep clustering algorithms on both image and mobile networks datasets (Sec. \ref{sec:eval}), as well as through an ablation study to examine the performance benefits of the different components (Sec. \ref{sec:abl}).
	\end{itemize}
	
	Our paper discusses two types of networks: neural networks, and mobile networks.
	To avoid confusion, we will refer to neural networks as "nets", while mobile networks will be referred to as "networks".

\section{State-of-the-Art in Deep Clustering}
	\label{sec:sota}


	For the purpose of this discussion, we can split deep clustering methods into two categories: \textit{generative} and \textit{discriminative}.
	Generative methods try to describe the data in a clustering-friendly representation.
	While learning, this representation is used both to define clusters, as well as to be able (re)construct data points into the original data space.
	Discriminative methods immediately try to divide the data into groups, without learning to recreate or generate data points in the process.

\begin{figure}[t]
		\centering
		\subfloat[Reconstructive]{
			\includegraphics[width=0.58\linewidth]{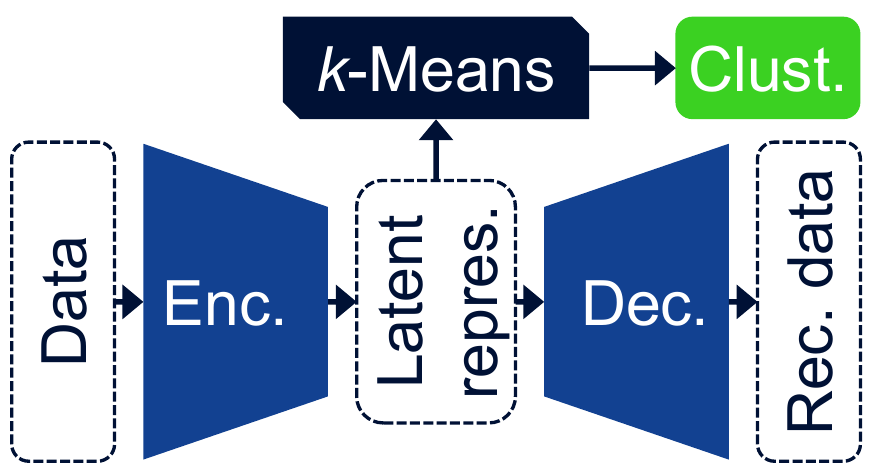}
			\label{fig:rec_topo}
		}
		\subfloat[Discriminative]{
			\includegraphics[width=0.36\linewidth]{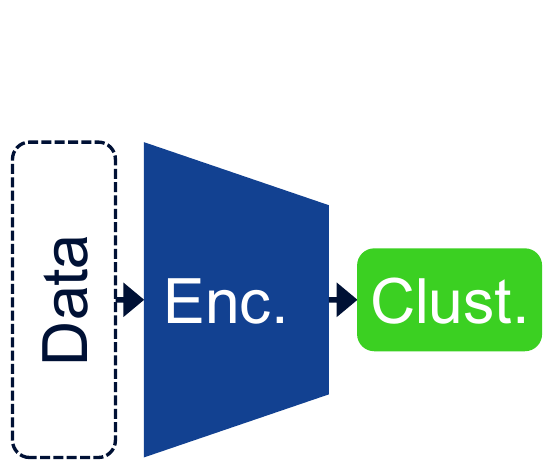}
			\label{fig:disc_topo}
		}
		\caption{The basic setup of the two main deep clustering approaches.}
		\label{fig:clust_topo}
	\end{figure}

	\textit{Reconstructive} methods, a subset of generative methods, learn to encode data into a simplified latent representation, from which the original data points can be decoded (reconstructed) effectively.
	For this purpose, most of the reconstructive clustering methods utilize autoencoder neural nets, made up of an encoder and a decoder sub-net (Fig. \ref{fig:rec_topo}). 
	By learning to distill information into a constrained latent space, autoencoders compress and reduce noise, formulating a high-level latent representation of the data, which contains only the most descriptive, meaningful features.
	Reconstructive clustering methods use these latent features on the assumption these are also the best descriptors of clusters in the data.
	
	One of the first examples of reconstructive deep clustering algorithms is \textit{\ac{DEC}} \cite{dec}.
	In \ac{DEC}, a stacked autoencoder is pre-trained, after which cluster centroids in the latent space are jointly optimized with the encoder, in order to best fit the encoded points to a predefined distribution around the centroids.
	This optimization causes the encoded points to tightly group around the cluster centroids, which has the effect of refining the clusters and increasing the nearest-neighbor assignment's accuracy. 
	We will discuss this algorithm in more detail in Section \ref{sec:rim_init_dec_clust}, as our proposed method is partly inspired by \ac{DEC}.
	Other similar methods, where the encoding is jointly optimized with the internal clustering are \ac{VaDE} \cite{vade} and \ac{DEPICT} \cite{depict}.
	
	\textit{\ac{ACAI}} \cite{acai} represents a different reconstructive approach, where the clustering does not influence the encoding.
	Instead, a separate mechanism is used to optimize the encoded representation for the later clustering step.
	\ac{ACAI} adopts an adversarial net commonly found in \acp{GAN} \cite{gan} to create believable data points when interpolating between encoded points in the latent space.
	Although not specifically meant for clustering, this regularization through believable interpolation leads to a highly clustering-friendly latent representation, where traditional clustering algorithm such as \kmeans{} \cite{kmeans} perform particularly well.
		
	Purely \textit{generative} methods are not as prevalent as reconstructive methods.
	One generative example is \ac{ClusterGAN} \cite{clustergan}, where a \ac{GAN} generator is used to synthesize believable data points from a mixture of categorical and continuous latent points.
	Apart from the usual \ac{GAN} setup of the generator (decoder) and adversarial nets, \ac{ClusterGAN} also implements an encoder, effectively realizing an inside-out autoencoder.
	Because purely generative methods seldom exist, we will sometimes refer to reconstructive methods as generative in this paper.
	
	Reconstructive methods assume that latent features learned through reconstruction are useful for clustering. Unfortunately this assumption does not hold for data in which clustering-irrelevant information (e.g.: small details, or information from other entities) outweighs clustering-relevant information.
	The best example of this can be seen in photographic datasets, where generative clustering algorithms often produce an effect labeled the \textit{blue sky problem}; planes, birds and other flying objects are all assigned to the same category, because the largest area of the image is taken up not by the object itself, but by the sky in the background.
	For the reconstruction of these images, the autoencoder pays more attention to correctly encode the sky, while losing sight of the clustering-relevant information about the objects in the latent representation.
	
	In another light, learning to decode (reconstruct) data serves as a regularizer in the formulation of the high-level latent representation.
	\textit{Discriminative} methods do away with this generative regularization, and replace it with their own specific regularization terms.
	This approach can have two benefits: the high-level representation can disregard information which is only useful for reconstruction, and the method can output cluster assignments directly, without the need for an additional step.
	Because of these advantages, discriminative clustering methods generally achieve higher accuracy while being more consistent on image datasets compared to their generative counterparts.
	The change in approach is also visible in the neural net topologies of these methods, usually consisting of a single sub-net, which effectively only implements the encoder half of an autoencoder (Fig. \ref{fig:disc_topo}).
	
	\textit{\ac{IMSAT}} \cite{imsat} was one of the first discriminative deep clustering methods to be published.
	\ac{IMSAT} builds on \ac{RIM} \cite{rim}, a (shallow) discriminative clustering approach which uses mutual information as a metric to develop cluster boundaries.
	In \ac{IMSAT}, the added \ac{SAT} procedure regularizes the encoding developed by \ac{RIM} so that the method can utilize deeper neural nets as encoder, without easily falling for degenerate models, thus arriving at better clustering accuracy on complex datasets.
	In a sense, SAT replaces the generative regularization.
	Another early discriminative method is \ac{DAC} \cite{dac}.
	
	The recently published \textit{\ac{DCCS}} \cite{dccs} method is especially interesting to our discussion.
	In \ac{DCCS}, the latent space is split into two feature groups, which the authors call category and style features.
	While all the latent features are used for information maximization, only the category features are used for clustering, which allows the method to further disregard irrelevant information, achieving even purer clusters and thus higher clustering accuracy.
	The regularization in \ac{DCCS} is done through a combination of adversarial nets and data augmentation in the form of randomized image transformations, such as cropping, aspect-ratio changes, hue and brightness changes, and the occasional horizontal flipping of the image.
	Other noteworthy discriminative deep clustering methods which use adversarial nets or data augmentation as regularization are \ac{ADC} \cite{adc} and \ac{IIC} \cite{iic}.
	
	Lastly, an often cited deep clustering method that does not fit into our categorization is \ac{JULE} \cite{jule}.
	\ac{JULE} is an agglomerative clustering algorithm, which creates clusters by merging individual observations, then smaller clusters, into ever bigger clusters.
	It is fundamentally different in its approach to the previously discussed methods here, which all at some point divide spaces or observations into clusters.
	
	Most of the above mentioned methods are developed for\nobreakdash-, and evaluated/benchmarked on image datasets.
	Commonly used image datasets for the evaluation of these algorithms are the MNIST\footnote{\url{http://yann.lecun.com/exdb/mnist/}} dataset containing $28\times28$ pixel greyscale images of handwritten digits, or the CIFAR-10\footnote{\url{https://www.cs.toronto.edu/~kriz/cifar.html}} dataset containing $32\times32$ pixel color photos of $10$ object categories (airplanes, cars, etc.).
	The clustering methods are trained in an unsupervised manner, without the input of the category labels, but are evaluated using the labels as ground truth.
	Their performance is measured using permutation-invariant external metrics, such as \ac{ACC}, \ac{NMI} or the \ac{ARI}, which quantify the similarity between the true category labels and the learned cluster assignments.
	Table \ref{tab:sota_perf} shows the published performance of the above discussed algorithms on the aforementioned (image) datasets.
	For the photographic CIFAR-10 dataset, many generative methods have no published results, and the ones that do, show worse performance than their discriminative peers, stemming from the previously discussed blue sky problem.
	
	\begin{table}[t]
		\centering
		\renewcommand{\arraystretch}{1.25}
		\setlength\tabcolsep{6pt}
		\caption{Performance of the state-of-the-art algorithms on the mnist and cifar-10 datasets. Values marked with * are taken from \cite{dccs}. All other values stem from the respective publications.}
		\label{tab:sota_perf}
		\begin{tabular}{l r| S S | S S }
			\multicolumn{2}{c|}{} & \multicolumn{2}{c|}{MNIST} & \multicolumn{2}{c}{CIFAR-10} \\
			\cline{3-6}
			Alg. & Year & ACC & NMI & ACC & NMI \\
			\hline
			\ac{DEC} \cite{dec}               & 2016 & 0.843 & 0.772{$^{*}$} & 0.301{$^{*}$}    & 0.257{$^{*}$} \\
			\ac{VaDE} \cite{vade}             & 2016 & 0.945 & 0.876{$^{*}$} & {$-$}            & {$-$}         \\
			\ac{DEPICT} \cite{depict}         & 2017 & 0.965 & 0.917         & {$-$}            & {$-$}         \\			
			\ac{ACAI} \cite{acai}             & 2019 & 0.962 & {$-$}         & {$-$}            & {$-$}         \\
			\ac{ClusterGAN} \cite{clustergan} & 2018 & 0.950 & 0.890         & {$-$}            & {$-$}         \\
			\hline
			\ac{IMSAT} \cite{imsat}           & 2017 & 0.984 & 0.956{$^{*}$} & 0.456            & {$-$}         \\
			\ac{DAC} \cite{dac}               & 2017 & 0.978 & 0.935         & 0.522            & 0.396         \\
			\ac{DCCS} \cite {dccs}            & 2020 & 0.989 & 0.970         & 0.656            & 0.569         \\			
			\ac{ADC} \cite{adc}               & 2019 & 0.987 & {$-$}         & 0.293            & {$-$}         \\
			\ac{IIC} \cite{iic}               & 2019 & 0.984 & 0.978{$^{*}$} & 0.576            & 0.513{$^{*}$} \\
			\hline
			\ac{JULE} \cite{jule}             & 2016 & 0.964 & 0.913         & 0.272{$^{*}$}    & 0.192{$^{*}$} \\
		\end{tabular}
	\end{table}

\section{Clustering with Decorrelating Adversarial Nets}
	\label{sec:dance}
	
	\begin{figure*}[t]
		\centering
		\includegraphics[width=\linewidth]{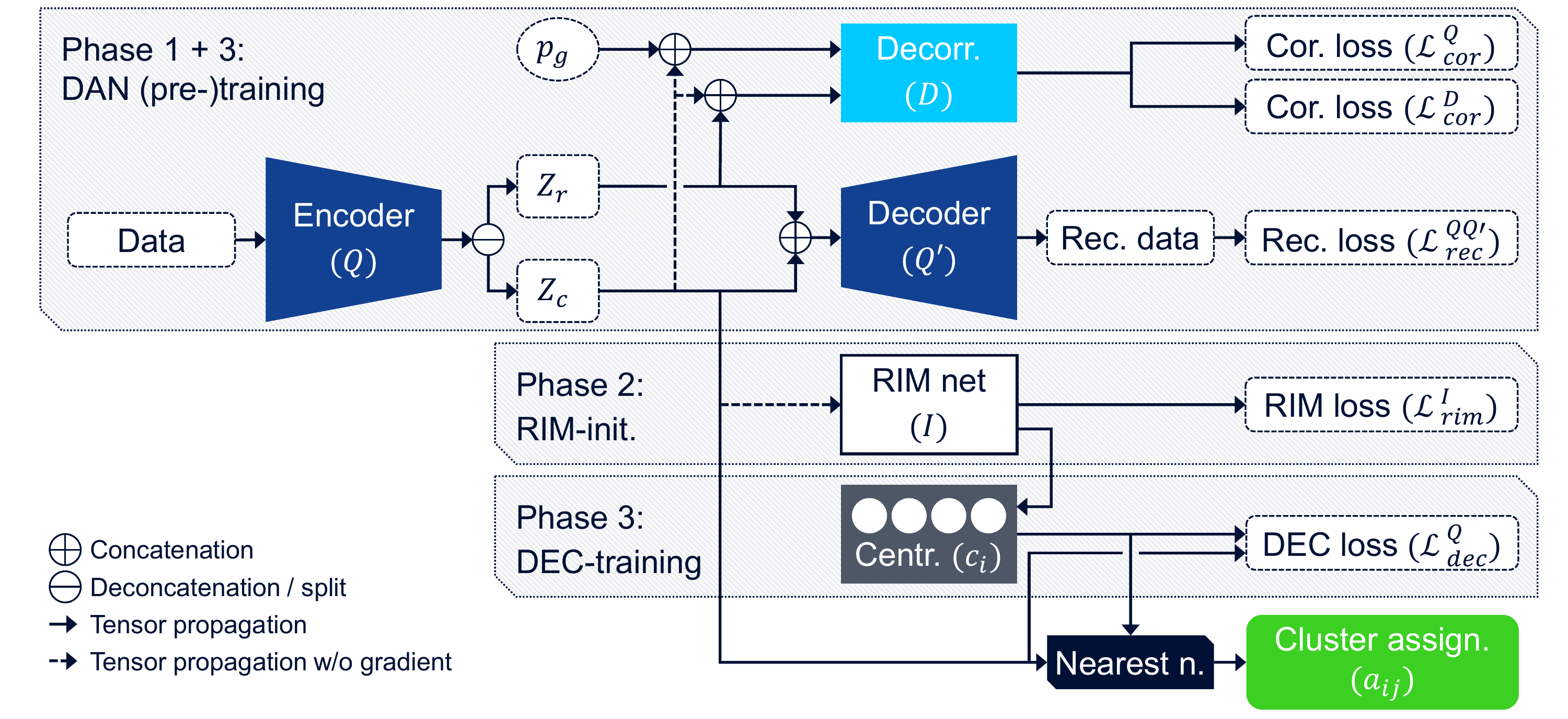}
		\caption{Overview of the components, training phases and losses in DANCE.}
		\label{fig:dan_overview}
	\end{figure*}

	\subsection{An Argument for Generative Clustering}
		\label{sec:gen_clust_arg}
		
		In the previous section, we discussed the blue sky problem, and it's detrimental effect on generative (reconstructive) clustering methods when used on image datasets.
		Although common in photographic data, specifically this kind of data pollution, i.e.: information from other entities seeping into the observations, is less prevalent in data from other domains.
		In mobile networks, data isolation is a given; for example, values logged about a cell in the network will never accidentally contain information about other cells.
		For this reason, discriminative methods might not see such a large advantage on mobile network data as they do on images when compared to generative algorithms.
		
		Furthermore, applying discriminative algorithms developed for image datasets to non-image datasets is not straight-forward.
		The specific regularization methods used by the discriminative approaches work well on image datasets, because biologically we humans have a great intuitive understanding of how vision and images behave, thus the authors were able to define meaningful augmentations of the data, which the methods then have to take into account in their model.
		The same can usually not be said about data from other domains: for example, image augmentations such as rotation or hue changes either don't make sense, or it is questionable if such variations are truly present in the data.
		In contrast, reconstructive methods don't suffer from such applicability problems; reconstruction always makes sense regardless of the data domain.
		
		However, all the above does not mean that generative clustering algorithms always perform better than discriminative algorithms on non-image datasets.
		In fact, as we will see later, generative algorithms often still show lower performance.
		Our observation while working with these algorithms is that a major reason for the worse average performance is the inconsistency of the internal clustering step.		
		The core of the reconstructive clustering methods is the training of the autoencoder, with clusters usually only defined in a subsequent step, sometimes not influencing the encoded features at all.
		The inconsistencies are a result of the applied simpler, ”traditional” clustering algorithms, such as \kmeans{} or Gaussian mixtures \cite{gmm}, which are sensitive to initialization and can get stuck in local minima when starting from an unfortunate position.		
		Some deep clustering methods that apply these simple clustering algorithms rectify/minimize this limitation by utilizing multiple initializations of the internal clustering step, trying to select the best clustering based on some unsupervised metric before proceeding.
		However, traditional (non-deep) clustering algorithms have no measures to effectively quantify the goodness of the	fit.
		
		We set out to create a method, which tries to improve on the above areas of generative clustering algorithms, with the objective of being consistent and easily applicable to non-image datasets, especially to data from mobile networks.
		What we propose here is not an entirely new algorithm, but an inventive way of using already established neural net components and training methods.
		The main aspects of our proposed method are:
		\begin{itemize}
			\item Easy cross-domain application stemming from the generative nature.
			\item Mitigation of the detrimental effects of reconstructive information perturbing the clustering features.
			\item A good initialization for the internal clustering.
		\end{itemize}	
	
	\subsection{Decorrelating Adversarial Net}
	
		Our proposal is called \ac{DANCE}, whose components and losses are illustrated by Fig. \ref{fig:dan_overview}.
		This section details the core of our proposed approach, the \ac{DAN}.
		
		\ac{DANCE} is based on an autoencoder neural net, made up of an encoder $Q$ and a decoder $Q'$.
		$Q$ realizes the non-linear mapping $Q(\theta_q, x) : X \rightarrow Z$, where $\theta_q$ are learnable parameters, and $Z$ is the encoded, latent feature space with a (much) lower dimensionality than the input (data) feature space $X$.
		$Q'(\theta_{q'}, z) : Z \rightarrow X$ approximates the inverse of $Q$, trying to reconstruct the original observations from $Z$.
		Both $\theta_q$ and $\theta_{q'}$ parameters are optimized through stochastic gradient descent, with the objective of minimizing the reconstruction loss:
		\begin{equation}
			\label{eq:l_qq_rec}
			\mathcal{L}^{QQ'}_{rec} = \frac{1}{n} \frac{1}{f} \sum_{i=1}^{n} \sum_{j=1}^{f} (x_{ij} - x_{ij}')^2,
		\end{equation}
		\noindent where $x$ and $x'$ denote the original and reconstructed data points, $f$ the number of features in $X$, and $n$ the number of data points in a batch, realizing the \ac{MSE} commonly used for regression tasks.		
		
		To reduce unnecessary reconstructive information used for clustering, we split $Z$ into two sets: features $Z_c$ that contain the clustering-relevant information, and features $Z_r$ that are purely reconstructive.
		We defined the following rules to distinguish the two sets:
		\begin{itemize}
			\item As features in $Z_r$ contain no clustering-relevant information, these must not be correlated to $Z_c$.
			\item As features in $Z_r$ contain only generic information which is applicable to all clusters, or include reconstruction-specific information about smaller, finer details in the data, $Z_r$ is likely to have a simple, noise-like distribution, such as a Gaussian distribution.
		\end{itemize}	
		
		To separate $Z_r$ from $Z_c$, the above description is posed as an adversarial game.
		Let $p_g$ refer to points sampled randomly from a Gaussian distribution with the same dimensionality as $Z_r$, $0$ mean and variance $\sigma$.
		To create non-correlated reference points with the desired distribution, we replace $Z_r$ in the original encoded features $Z = Z_c \oplus Z_r$ by $p_g$, arriving at $Z' = Z_c \oplus p_g$ ($\oplus$ denotes concatenation).
		An adversarial net $D$ (decorrelator) has the task to detect if a point comes from $Z$ or $Z'$, by formulating rules that either consider the difference in distribution between $Z_r$ and $p_g$, or detect correlation between $Z_r$ and $Z_c$.
		The encoder $Q$ has to generate a latent encoding $Z$ which is impossible to differentiate from $Z'$, thus mimicking the distribution of $p_g$ with $Z_r$ and breaking any correlation between $Z_c$ and $Z_r$.
		
		$D(\theta_{d}, z): Z \rightarrow d:[0, 1]$ outputs a singular scalar which represents the estimated probability that $z$ came from $Z$ rather than $Z'$.
		To optimize the decorrelator and the encoder parameters for the adversarial game, stochastic gradient descent is used.
		The respective losses, which realize the minimization of the Jensen–Shannon divergence as proposed in the original \ac{GAN} paper \cite{gan} are the following:
		\begin{equation}
			\label{eq:l_q_cor}
			\mathcal{L}^Q_{cor} = -\frac{1}{n} \sum_{i=1}^{n} log(d_i),
		\end{equation}			
		\begin{equation}
			\label{eq:l_d_cor}
			\mathcal{L}^D_{cor} = -\frac{1}{n} \sum_{i=1}^{n} log(d'_i) + log(1 - d_i),
		\end{equation}
		\noindent where $d_i$ and $d'_i$ refer to decorrelator guesses on points coming from $Z$ and $Z'$, and $n$ refers to the number of data points in a batch.		
		
		The above described neural net setup is very similar to Wasserstein Autoencoders \cite{wae}.
		As such, in theory the decorrelation does not interfere with the original autoencoding task, and any desired loss function could work well to calculate the reconstruction loss.
		To balance the losses that affect the encoder, coefficient $\beta_{cor}$ can be introduced, so that the final encoder loss is:
		\begin{equation}
			\label{eq:l_q}
			\mathcal{L}^Q = \mathcal{L}^{QQ'}_{rec} + \beta_{cor}\mathcal{L}^Q_{cor}
		\end{equation}
		
		A major problem with the decorrelator in this format is the continuous disturbance of the features in $Z_c$ during training.
		As the correlation between $Z_c$ and $Z_r$ can be broken in both feature sets, $\mathcal{L}^Q_{cor}$ generates gradients which try to move the points around in a chaotic manner in $Z_c$.
		Since the adversarial game does not define any target distribution for $Z_c$, this disturbance does not converge, and is constantly present during training, in the worst cases causing $Z_c$ to collapse into a single point.
		An obvious choice would be to impose a prior distribution on $Z_c$, the same way we are already imposing a prior on $Z_r$ (similarly how to some extent \ac{DCCS} does), however, in our experience priors on $Z_c$ don't work well in the autoencoder setting.
		Instead, in \ac{DANCE} backpropagation of the gradient from $\mathcal{L}^Q_{cor}$ is stopped through $Z_c$, so that the adversarial game does not affect those features.
		This effect can be achieved in most deep learning frameworks with a simple detach() or stop\textunderscore gradient() call.
		This gradient stop still allows the $Z_c$ features to be used by $D$ for the detection of correlation with $Z_r$, but the gradients only affect $Z_r$, leaving $Z_c$ undisturbed by the decorrelation.
		
\begin{figure}[t]
			\centering
			\subfloat[$Z_r$ density]{
				\includegraphics[width=0.45\linewidth]{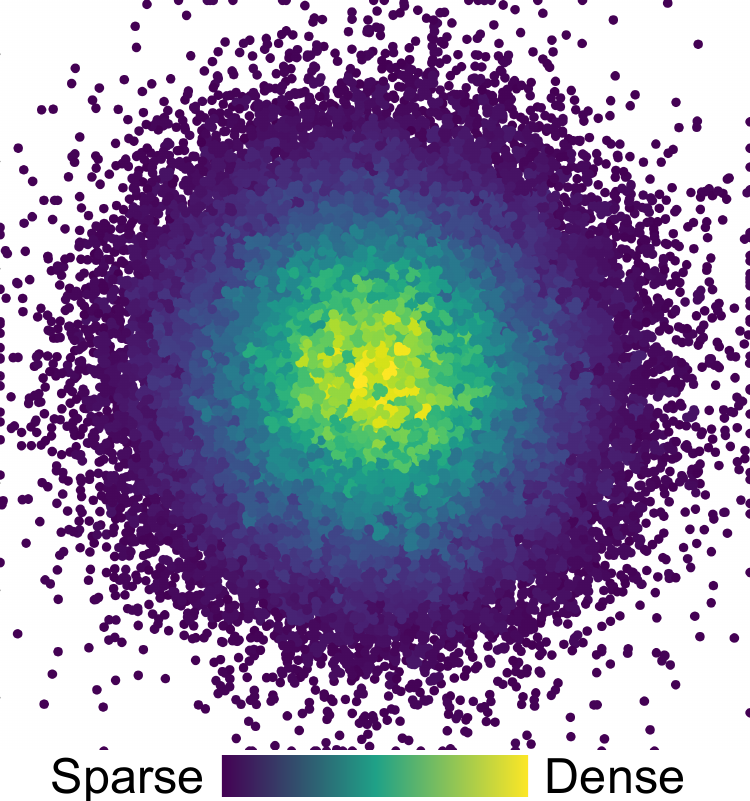}
				\label{fig:enc_mnist_zr_den}
			}
			\subfloat[$Z_r$ ground truth]{
				\includegraphics[width=0.45\linewidth]{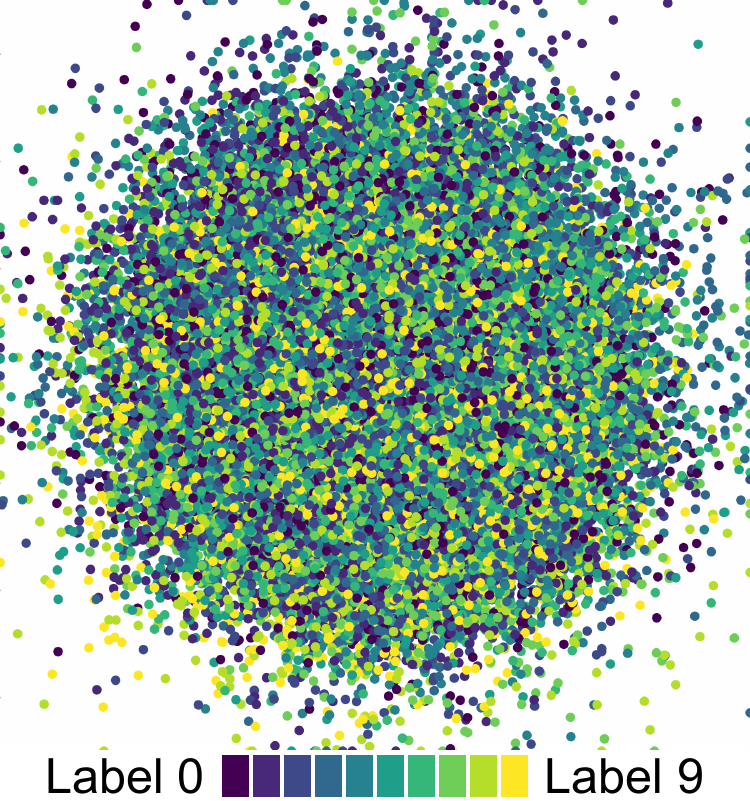}
				\label{fig:enc_mnist_zr_lab}
			}\\
			\subfloat[$Z_c$ density]{
				\includegraphics[width=0.45\linewidth]{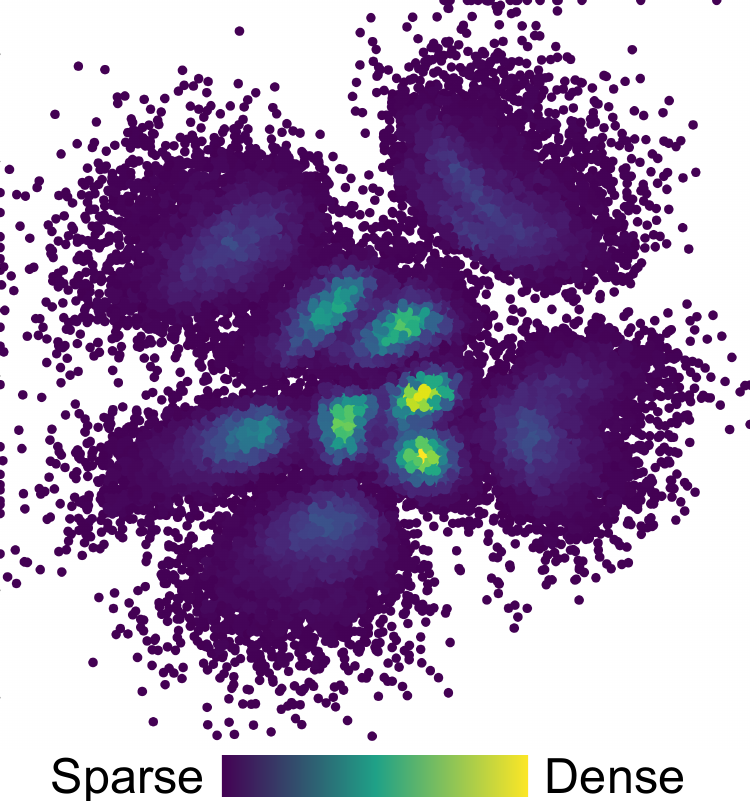}
				\label{fig:enc_mnist_zc_den}
			}
			\subfloat[$Z_c$ ground truth]{
				\includegraphics[width=0.45\linewidth]{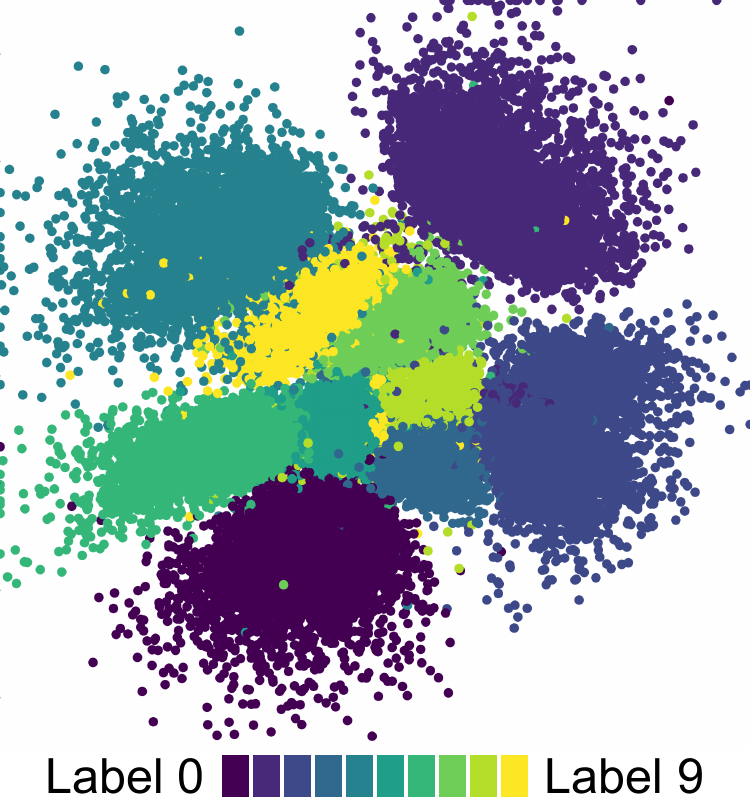}
				\label{fig:enc_mnist_zc_lab}
			}
			\caption{A typical \ac{DANCE} encoding of the MNIST dataset at the end of the \ac{DAN} pre-training. The top two figures depict how well $Z_r$ follows the $p_g$ Gaussian prior, as well as being completely decorrelated to $Z_c$. The bottom two figures show the irregular, but well separated encoding in $Z_c$.}
			\label{fig:enc_mnist}
		\end{figure}

		In theory, the \ac{DANCE} setup could lead to an encoding where both clustering and reconstructive information is communicated only through $Z_c$, and $Z_r$ does not carry any information at all, only capturing random noise in order to adhere to the $p_g$ prior.
		In our experience this is never the case, as the autoencoder always tries to utilize all latent features. 
		However, possibly for this reason, or because of saturation problems in $D$, \ac{DANCE} seems to work best if both $Z_c$ and $Z_r$ are of low dimensionality.
		On Figure \ref{fig:enc_mnist}, a typical \ac{DANCE} encoding of the MNIST dataset can be seen, where both $Z_c$ and $Z_r$ are $2$-dimensional.
		
		We realize that the above detailed decorrelation does not guarantee that all clustering-relevant information ends up in $Z_c$, nor that all clustering-irrelevant information is removed from $Z_c$.
		In reality, \ac{DANCE} reduces variance in $Z_c$, and allows for a more coherent mapping where similar data points are close together, which is beneficial for a subsequent application of traditional clustering algorithms.
	
	\subsection{RIM Initialization and DEC Clustering}
		\label{sec:rim_init_dec_clust}
	
		The internal clustering step in \ac{DANCE} is done with the mechanism from \ac{DEC}.
		In its original form, \ac{DEC} uses the \kmeans{} algorithm to find initial positions for the cluster centroids.
		We found this initialization to be quite unreliable, because \kmeans{} is biased towards convex, even-sized clusters by design, which is often not how the encoded clusters behave in \ac{DANCE}.
		Instead, we opted to use the discriminative \ac{RIM} \cite{rim} algorithm to find a good initial clustering, which is then subsequently refined by \ac{DEC}.
		
		In \ac{RIM}, a simple feed-forward neural net is used to find clusters, by looking for cluster boundaries that are in sparsely populated regions of the input data space.
		To achieve this, \ac{RIM} minimizes the conditional entropy, balanced by the maximization of the entropy of the label distribution, which helps to form clusters with even populations.
		In effect, this optimization task maximizes the empirical estimate of the mutual information between data point and their assignment.
		Let $I(\theta_i, z_c) : Z_c \rightarrow P$ denote the \ac{RIM} net, where $\theta_i$ are learnable parameters.
		$I$ directly outputs cluster assignment probabilities for each input point.
		To train the \ac{RIM} net, the following loss terms are minimized through stochastic gradient descent:
		\begin{equation}
			\label{eq:l_i_Cent}
			\mathcal{L}^I_{cond.ent} = -\frac{1}{n} \sum_{i=1}^{n} \sum_{j=1}^{k} p_{ij}log(p_{ij}),
		\end{equation}
		\begin{equation}
			\label{eq:l_i_Lent}
			\mathcal{L}^I_{lab.ent} = \sum_{j=1}^{k} ( \frac{1}{n} \sum_{i=1}^{n} p_{ij} ) \times log(\frac{1}{n} \sum_{i=1}^{n} p_{ij}),
		\end{equation}
		\begin{equation}
			\label{eq:l_i_rim}
			\mathcal{L}^I_{rim} = \mathcal{L}^I_{cond.ent} + \mu\mathcal{L}^I_{lab.ent} + \lambda R(\theta_i),
		\end{equation}
		\noindent where $p_{ij}$ refers to the cluster assignments output by $I(\theta_i, z_c)$, $k$ refers to the number of clusters, $\mu$ is a balancing coefficient between the two entropy terms, and $\lambda$ is a balancing coefficient for the regularization term $R(\theta_i)$.		
		For the $R(\theta_i)$ regularization term we used the $L_2$ norm of the parameters $\theta_i$, also commonly referred to as weight decay.
		
\begin{figure}[t]
			\centering
			\subfloat[\ac{RIM} initialization]{
				\includegraphics[width=0.45\linewidth]{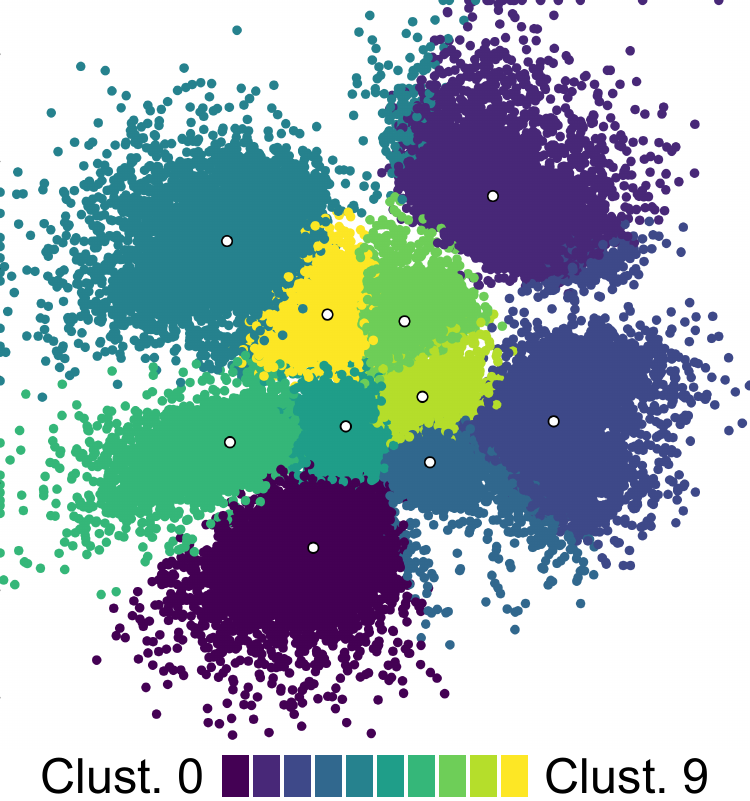}
				\label{fig:clust_mnist_rim_ass}
			}
			\subfloat[\ac{DEC} refinement]{
				\includegraphics[width=0.45\linewidth]{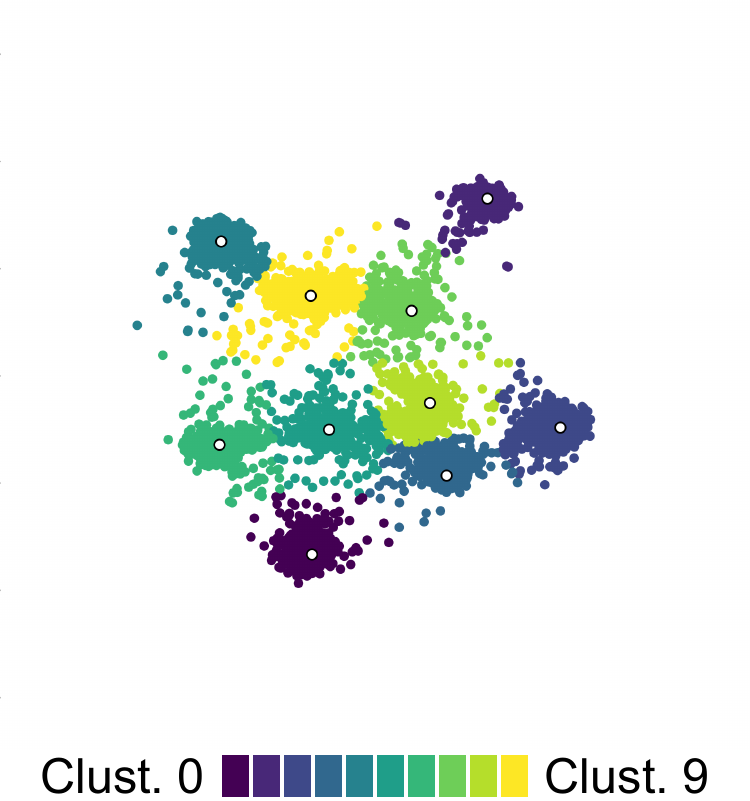}
				\label{fig:clust_mnist_dec_ass}
			}
			\caption{RIM initialization and DEC refinement on the encoded $Z_c$ (MNIST). The white dots represent the cluster averages (centroids), the coloration shows the cluster assignments.}
			\label{fig:clust_mnist}
		\end{figure}

		We found that the achieved conditional entropy at the end of the \ac{RIM} training is also a somewhat good indicator of the objective goodness of the clustering, rectifying a problem mentioned in Sec. \ref{sec:gen_clust_arg}.
		To exploit this feature, we re-train $I$ multiple times for a fixed number of epochs, and select the training with the lowest $\mathcal{L}^I_{Cent}$ at the end.
		The averages of the clusters in $Z_c$ found by this $I$ are then used as centroids to start the \ac{DEC} refinement.
		Figure \ref{fig:clust_mnist_rim_ass} shows the clusters and the subsequent cluster centroids defined by \ac{RIM} on the \ac{DANCE} encoded MNIST dataset.
		
		\ac{DEC} uses the initialized cluster centroids to influence the latent space in $Z_c$, by moving the encoded points closer to their assigned centroids, while simultaneously moving the centroids to best fit their assigned points \cite{dec}.
		To do this, \ac{DEC} uses Student's $t$-distribution to define a soft-assignment between encoded points and cluster centroids.		
		The encoded points and the centroids are then jointly moved to best fit a target distribution by minimizing the Kullback-Leibler divergence.
		Because of the soft assignment, the \ac{DEC} optimization has a more pronounced effect on the encoded points closest to the centroids compared to more remote points, and vice versa; the centroids are moved to best fit the closest encoded points.
		The \ac{DEC} loss is defined as:
		\begin{equation}
			q_{ij} = -\frac{ (1 + \|z_{c_i} - c_j\|^2/\alpha)^{-\frac{\alpha + 1}{2}} }{ \sum_{j'=1}^{k} (1 + \|z_{c_i} - c_j'\|^2/\alpha)^{-\frac{\alpha + 1}{2}} },
		\end{equation}
		\begin{equation}
			p_{ij} = -\frac{ q_{ij}^2 / \sum_{i=1}^{n} q_{ij} }{ \sum_{j'=1}^{k} (q_{ij'}^2 / \sum_{i=1}^{n} q_{ij'}) },
		\end{equation}
		\begin{equation}
			\label{eq:l_q_dec}
			\mathcal{L}^Q_{dec} = \sum_{i=1}^{n} \sum_{j=1}^{k} p_{ij}log \frac{p_{ij}}{q_{ij}},
		\end{equation}
		\noindent where $c_j$ refers to cluster centroids, $\alpha$ is the degrees of freedom of the Student’s $t$-distribution, $q_{ij}$ is the soft assignment of the encoded points, and $p_{ij}$ is the auxiliary distribution.
		During the \ac{DEC} clustering phase, the encoder loss expands to:
		\begin{equation}
			\mathcal{L}^Q = \mathcal{L}^{QQ'}_{rec} + \beta_{cor}\mathcal{L}^Q_{cor} + \beta_{dec}\mathcal{L}^Q_{dec},
		\end{equation}
		\noindent where $\beta_{dec}$ is an additional coefficient balancing the \ac{DEC} loss with the others.
		
		The \ac{DEC} loss tightly groups the encoded points around the centroids (Fig. \ref{fig:clust_mnist_dec_ass}), which further refines the latent space by forcing the autoencoder to make "decisions" about where encoded points end up.
		This tight grouping also compensates for the limitation of nearest-neighbor clustering, because the tight groups can be efficiently separated by linear boundaries (Voronoi tessellation).
		Without this grouping, straight cluster boundaries often don't align with the correct distribution-boundaries, making nearest-neighbor clustering algorithms, especially \kmeans{}, ineffective in these cases.

		\begin{algorithm}[ht]
			\label{alg:dance}
			
			\SetAlgoLined
			\newcommand{\nosemic}{\SetEndCharOfAlgoLine{\relax}}   
			\newcommand{\dosemic}{\SetEndCharOfAlgoLine{\string;}} 
			\newcommand{\pushline}{\Indp}                          
			\newcommand{\popline}{\Indm\dosemic}                   
								
			\SetKwInput{Kw}{Input}			
			\KwIn{Dataset $\mathcal{X} = \{x_i\}^N_{i=1}$, initial parameters $\theta_q$, $\theta_{q'}$, $\theta_d$, $\theta_i$ of encoder $Q$, decoder $Q'$, decorrelator $D$ and \ac{RIM} initializer $I$, nr. of dimensions $n_{Z_c}$, hyper-parameters $\sigma$, $\alpha$, $\mu$, coefficients $\beta_{cor}$, $\beta_{dec}$.}
			
			\SetKwFunction{encode}{autoencode}
			\SetKwProg{proc}{Procedure}{}{}
			\SetKwProg{pretrain}{DAN pre-training}{}{}
			\SetKwProg{riminit}{RIM initialization}{}{}
			\SetKwProg{dectrain}{DEC refinement}{}{}
  			
			\proc{\encode{}}{
				\nosemic Compute $z = Q(\theta_q, x)$, $x' = Q'(\theta_{q'}, z)$ and using\;
				\pushline\dosemic these $\mathcal{L}^{QQ'}_{rec}$ (Eq. \ref{eq:l_qq_rec})\;
				
				\popline Generate $z' = z_c \oplus p_g(\sigma)$\;
				
				\nosemic Compute $d = D(\theta_{d}, z)$, $d' = D(\theta_{d}, z')$ and using\;
				\pushline\dosemic these $\mathcal{L}^Q_{cor}$ (Eq. \ref{eq:l_q_cor}) and $\mathcal{L}^D_{cor}$ (Eq. \ref{eq:l_d_cor})\;
			}
		
			\pretrain{}{
				\For{$e_{pre}$ iterations}{
					\encode{}\;
					Update $\theta_d$ by minimizing $\mathcal{L}^D_{cor}$\;
					\nosemic Update $\theta_q$, $\theta_{q'}$ by minimizing\;
					\pushline\dosemic $\mathcal{L}^Q = \mathcal{L}^{QQ'}_{rec} + \beta_{cor}\mathcal{L}^Q_{cor}$ (Eq. \ref{eq:l_q})\;	
				}
			}

			\riminit{}{
				Deconcatenate $z_c = Q(\theta_q, x) \ominus(n_{Z_c})$\;
				\For{$n_{rim}$ tries}{
					\For{$e_{rim}$ iterations}{
						\nosemic Compute $p = I(\theta_i, z_c)$ and using\;
						\pushline\dosemic it $\mathcal{L}^I_{rim}$ (Eq. \ref{eq:l_i_rim})\;
						\popline Update $\theta_i$ by minimizing $\mathcal{L}^I_{rim}$\;
					}
					Compute $z = I(\theta_i, z_c)$ and using it $\mathcal{L}^I_{Cent}$ (Eq. \ref{eq:l_i_Cent})\;
					Store $\theta_{i_{best}} = \theta_i$ if $\mathcal{L}^I_{Cent}$ is the lowest so far\;
				}
				Compute $a_{ij} = argmax(I(\theta_{i_{best}}, z_c))$ assignments\;
				\For{$j = 1...k$}{
					$c_j = {\sum_{i=1}^{n} z_{c_i}(a_{ij} == j)} / {\sum_{i=1}^{n} (a_{ij} == j)}$\;
				}
			}
		
			\dectrain{}{
				\For{$e_{dec}$ iterations}{
					\encode{}\;
					Compute $\mathcal{L}^Q_{dec}$ (Eq. \ref{eq:l_q_dec}) using $z_c$, $c$\;
					Update $\theta_d$ by minimizing $\mathcal{L}^D_{cor}$\;
					\nosemic Update $\theta_q$, $\theta_{q'}$, $\mu$ by minimizing\;
					\pushline\dosemic $\mathcal{L}^Q = \mathcal{L}^{QQ'}_{rec} + \beta_{cor}\mathcal{L}^Q_{cor} + \beta_{dec}\mathcal{L}^Q_{dec}$				
				}
			}
			Compute $a_{i} = argmin((c_j - z_{c_i})^2 )$ final assignments\;
			\caption{\ac{DANCE}}
		\end{algorithm}
	
		The complete \ac{DANCE} algorithm is shown by Alg. \ref{alg:dance}.
		The three main phases: \ac{DAN} pre-training, \ac{RIM} initialization and \ac{DEC} refinement can also be seen as overhauled versions of the original \ac{DEC} training steps, but we hope the reader agrees that the changes are significant enough to warrant a different algorithm name.
		Although other papers often refer to multi-phase methods in a negative light, we have found that the isolated phases are easier to debug or parameterize, not having to completely restart training if something is not perfect, which also helped us during the evaluation of the algorithm.

\section{Evaluation}
	\label{sec:eval}

	\subsection{Methodology}
		
		Our ultimate goal was to evaluate \ac{DANCE} on the mobile network dataset, but in order to be able to compare the performance to other state-of-the-art methods with this dataset, we needed to have working implementations of the other algorithms.
		As it would be an enormous undertaking to try to re-implement and evaluate every method we listed in Section \ref{sec:sota}, we have selected four methods to compare against, based on their performance, age, and their connection to our algorithm:
		\begin{itemize}
			\item \textbf{\ac{DEC}} is an obvious choice for an older generative algorithm, as \ac{DANCE} shares its internal clustering and the overall structure.
			In both methods, the internal clustering influences the encoding.
			\item \textbf{\ac{ACAI}} is a state-of-the-art generative clustering algorithm, which, contrary \ac{DEC} and \ac{DANCE}, develops the encoding independent from the internal clustering.
			\item \textbf{\ac{IMSAT}} is one of the first discriminative algorithms, working without manual data augmentation for the regularization. \ac{IMSAT} builds on \ac{RIM}, which we also utilize in \ac{DANCE}.
			\item \textbf{\ac{DCCS}} is our choice of a state-of-the-art discriminative algorithm.
			In contrast to \ac{IMSAT}, \ac{DCCS} does use explicit data augmentation as regularization, and shares the core idea of separating clustering-relevant from clustering-irrelevant features with \ac{DANCE}.
		\end{itemize}
		\noindent Altogether, we have an even split of generative and discriminative deep clustering algorithms, as well as an even split between earlier and recent publications.
		This aspect is important, because we suspected that older algorithms are by no means necessarily worse than newer publications on mobile network data, contrary to what the trend might show on image datasets.
		
		Another beneficial effect of having to re-implement contending methods is that we can control for neural net complexity in the evaluation.
		Even in the few years since the first publication of these methods, the emergence of dedicated hardware accelerators, easy-to-use deep learning frameworks, and new type of components and training methods increased the possibility of training deep neural nets tremendously.
		While newer publications use residual nets tens or even hundreds of layers deep, some older algorithms were evaluated using only a few simple fully-connected layers.
		We suspected that the topological differences account for a large portion of the performance differences, and we were interested in understanding how the algorithms differ in performance when using the same neural net components and sizes.
		Thus, in the following evaluation, all methods use the same convolutional encoder, convolutional decoder and fully-connected adversarial nets (per dataset).
		
		We ran multiple ($8$) trainings of all methods for each dataset, to be able to present worst, average and peak performance metrics.
		Of course, not having invented and exhaustively fine-tuned these methods, we likely can not utilize them to their utmost potential, and probably leave a few percentage points of accuracy on the table.
		On the other hand, the usability and applicability of an algorithm is just as important as peak performance, and is a focus of this paper.
		We ask the reader to keep this in mind while reading the following sections.

	\subsection{Evaluation on Image Data}
		
		To give a performance comparison in a common setting, and to establish that our implementations are working correctly, we first evaluated the methods on the MNIST image dataset.
		This also allows us to compare \ac{DANCE} to other state-of-the-art algorithms' performance without having to implement them.
		Lastly, this evaluation gave us a chance to see if older algorithms show improved performance over their published results when using our deeper convolutional nets.
		The performance of the compared methods can be seen on Table \ref{tab:mnist_perf}.
		The (external) metrics we utilize throughout this evaluation are:
		\begin{itemize}
			\item \textbf{\ac{ACC}}, which measures the ratio between number of points correctly assigned against the number of all datapoints in the dataset.
			As the mapping between labels and clusters is ambiguous, we used the Hungarian method \cite{hungarian} to determine the best mapping/permutation, thus making the metric permutation-invariant.
			\item \textbf{\ac{NMI}}, which measures the mutual information between labels and cluster assignments.
			\ac{NMI} is normalized so that $0$ means no mutual information, while $1$ is the maximal mutual information achievable.
		\end{itemize}
	
		
		
		\begin{table}[h]
			\centering
			\renewcommand{\arraystretch}{1.25}
			\setlength\tabcolsep{4pt}
			\caption{Performance of the evaluated algorithms \\ on the mnist dataset}
			\label{tab:mnist_perf}
			\begin{tabular}{l|c|c|c}
				           & \multicolumn{2}{c|}{ACC}                     & NMI \\
				\cline{2-4}
				Alg.       & avg ($\pm$std)         & min - max           & avg ($\pm$std) \\
				\hline
				\ac{DEC}   & $0.9385$ ($\pm0.045$) & $0.8882$ - $0.9898$ & $0.9313$ ($\pm0.031$) \\
				\ac{ACAI}  & $0.9525$ ($\pm0.040$) & $0.8482$ - $0.9774$ & $0.9171$ ($\pm0.020$) \\
				\ac{IMSAT} & $\mathbf{0.9866}$ ($\pm0.004$) & $\mathbf{0.9776}$ - $\mathbf{0.9904}$ & $\mathbf{0.9629}$ ($\pm0.006$) \\
				\ac{DCCS}  & $0.9490$ ($\pm0.043$) & $0.8760$ - $0.9829$ & $0.9338$ ($\pm0.028$) \\
				\ac{DANCE} & $0.9625$ ($\pm0.016$) & $0.9368$ - $0.9806$ & $0.9249$ ($\pm0.019$) \\
			\end{tabular}
		\end{table}

		The two generative algorithms, \ac{DEC} and \ac{ACAI} exhibits large standard deviation in accuracy and mutual information, which we attribute to the inconsistency of the traditional clustering algorithms in this setting; \kmeans{} cannot reliably find the true clusters in the encoding, and often converges to local minima, arriving at sub-optimal fits.
		
		Apart from this, \ac{DEC} performed quite a lot better than the originally published results (shown in Table \ref{tab:sota_perf}) using our deeper encoder and decoder.
		It is especially important to note the impressive maximum accuracy, which is by far the best out of any reconstructive algorithm.
		Also quite interesting is the relatively high \ac{NMI} achieved compared to the \ac{ACC}, which represents a high mutual information content between ground truth labels and cluster assignments.
		This phenomenon occurs when wrongly clustered observations have a systematic error.
		In the MNIST dataset, an example of such a systematic error would be that the cluster containing all $9$s also includes a few $7$s, but not any other numbers.
		
		The \ac{ACAI} results are a little worse than the originally published results in Table \ref{tab:sota_perf}.
		This is not a mistake or misconfiguration on our part; in the original paper the authors themselves admit to selecting the best \kmeans{} clustering based on \textit{external} metrics, reasoning that the \ac{ACAI} algorithm was anyway not originally intended for clustering, and that the shown results are only there to signify the potential of such an approach \cite{acai}.
 		In order to provide a fair and unsupervised comparison of the algorithms, we selected the best out of repeated \kmeans{} fits using an \textit{internal} metric (not utilizing the ground truth) in our evaluation, hence the worse results.

		\ac{IMSAT} performed phenomenally, even improving on the already excellent originally published results and coming close to the published performance of \ac{DCCS}.
		Of note is the very low deviation, and high minimum values, which is a nice guarantee for the user that even in the worst case the clustering is almost the best it can be.
		This trust would be very important for unsupervised algorithms, as the user has no way of confirming the quality of the clustering in a real-life scenario.
		
		\ac{DCCS}, on the other hand, proved quite sensitive to the net topology, and performed much worse than the published performance in Table \ref{tab:sota_perf}.
		We were definitely able to reproduce the originally published results using the proposed net, but switching to our more complex topology caused \ac{DCCS} to learn sub-optimal fits, where often one or more of the clusters were left unused (unpopulated).
		We tried rectifying this through changing the balance of the prior loss, as well as adding batch-normalization layers to $Z_c$, but to no avail.
		It seems to us that the net complexity plays a major role for \ac{DCCS} in inherently regularizing the model, and more complex nets are not regularized sufficiently with only the additional mechanism in \ac{DCCS}.
		This is very counter-intuitive, as all the other algorithms benefited from the increased modeling capability of the more complex neural net used.
		
		Our \ac{DANCE} algorithm performed as expected; excellent for a generative clustering method, yet not in the range of most discriminative methods' capabilities.
		Compared to the published values shown in Table \ref{tab:sota_perf}, \ac{DANCE} is among the best performing generative approaches on the MNIST dataset.
		We are particularly happy with the quite high minimum accuracy metric, which again plays a major role in establishing trust towards the algorithm.
		As far as we can tell, most of the loss in accuracy stems from malformed latent representations, where parts of a class ends up separated, far away from most of the points in the same class.
		This effect could be possibly mitigated by utilizing more dimensions for $Z_c$, but in our experience the gain in consistency is counteracted by the loss in performance both from the decorrelation and from the \ac{RIM} initialization, negating any benefit.
		
		Lastly, our goal with this evaluation was to tune most of the hyper-parameters of the algorithms, and apart from the net topologies, reuse these settings in the mobile data evaluation.
		However, we realized that the difference between the two domains likely causes the hyper-parameters to be sub-optimal for the network dataset, so in the end we allowed the tuning of some parameters based on internal metrics, such as balancing losses, or adjusting learning rates.
		These changes could be reasonably made without the knowledge of the ground truth, in order to stay within the bounds of a realistic clustering scenario.
		Furthermore, every algorithm has seen increased training iterations to compensate for the smaller dataset, resulting in less updates per epoch.	
		
	\subsection{Evaluation on Mobile Network Data}
		
		\acused{VoIP}
		\acused{HTTP}
		\acused{FTP}
		
		\begin{table}[t]
			\centering
			\renewcommand*{\arraystretch}{1.25}
			\caption{User groups in the mobile network dataset.}
			\label{tab:user_groups}
			\begin{tabular}{l|c|c|c|c}
				    & Label        & Traffic   & Speed [km/h] & Movement \\
				\hline
				$0$ & Stationary 1 & \ac{FTP}  & $0$          & -       \\
				$1$ & Stationary 2 & \ac{VoIP} & $0$          & -       \\
				$2$ & Stationary 3 & \ac{HTTP} & $0$          & -       \\
				$3$ & Pedestrian 1 & \ac{FTP}  & $8$          & random  \\
				$4$ & Pedestrian 2 & \ac{VoIP} & $8$          & random  \\
				$5$ & Pedestrian 3 & \ac{HTTP} & $8$          & random  \\
				$6$ & Vehicular 1  & \ac{VoIP} & $10$ - $100$ & streets \\
				$7$ & Vehicular 2  & \ac{HTTP} & $10$ - $100$ & streets \\
			\end{tabular}
		\end{table}
				
		The evaluation on mobile network data represents the target clustering scenario for our algorithm, which also tests the other algorithms' flexibility towards different application domains.
		The evaluation we devised tries to pose a similar task to the common problem in the use-cases introduced in Sec. \ref{sec:intro}: the clustering of behavioral patterns.		
		In our evaluation, the clustering algorithms were to assign mobile users to groups based on how they use the network, and what they use it for, using information that implicitly describes their behavior.
		
\begin{figure}[h]
			\centering
			\includegraphics[width=\linewidth]{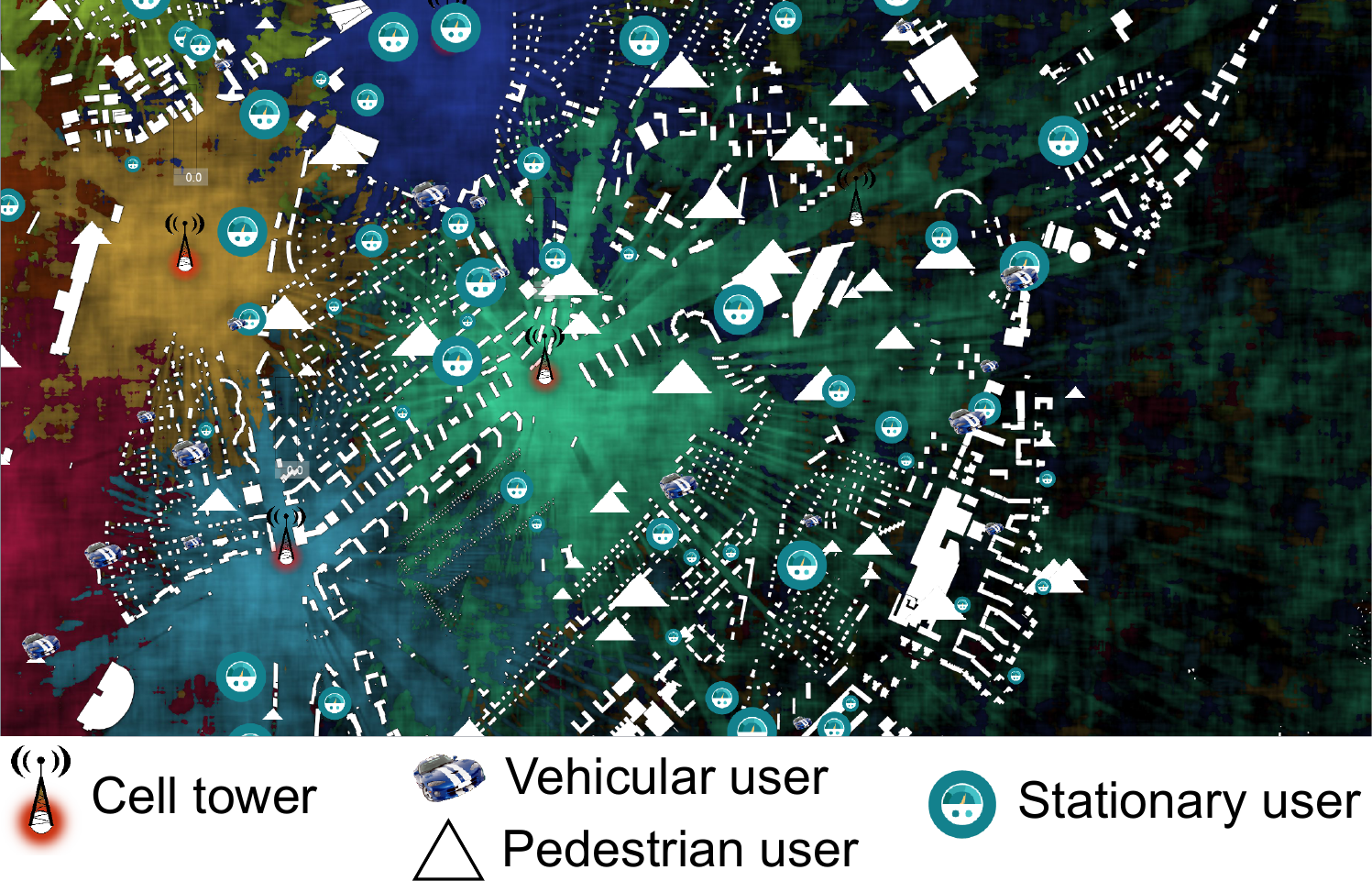}
			\caption{Excerpt from the Helsinki simulation scenario.}
			\label{fig:sim_scenario}
		\end{figure}

		In order to generate data where we know the ground truth, we used a mobile network simulator, which enables us to use the usual external metrics in our evaluation.
		The simulation scenario was set in the city of Helsinki, where mobile users moved around and used the multi-layer heterogeneous network to communicate (Fig. \ref{fig:sim_scenario}).
		The network comprised of multiple macro, micro and WiFi cells (access points), and covered most of the city.
		The users were allocated into $8$ user groups, which were differentiated based on the user's mobility patterns (stationary, pedestrian, vehicular) and their network usage type (talking using \ac{VoIP}, web-browsing using \ac{HTTP} and transferring files using \ac{FTP}).
		The definitions of the user groups can be seen in Table \ref{tab:user_groups}.
		
		The collected data contained:
		\begin{itemize}
			\item \textbf{Application} level \acp{KPI} such as downlink and uplink throughput.
			\item \textbf{Radio quality} indicators, such as \ac{CQI}, \ac{SNR}, scheduling delay and \ac{RSRP}.
			\item \textbf{\ac{RRC}} state indicators, such as connected, \ac{RLF} and idle states.
		\end{itemize}
		\noindent A total of $17$ values were collected every $5$ seconds for every user.
		The simulation contained $400$ users, an even distribution of $50$ users from each of the $8$ user groups.
		Each user was observed for $10$ consecutive sequences, with a sequence consisting of $256$ time steps
		In total this corresponds to about $3.5$ hours of simulation time.
		The collected data was organized into an array with the shape of $4000 \times 256 \times 17$, which is functionally the same as $256 \times 1$ pixel images containing $17$ channels (instead of the usual $3$: red, green and blue).
		The clustering algorithms processed this data using $1$-dimensional convolutional encoders (and decoders).
		The resulting performances can be seen on Table \ref{tab:mobile_perf}.
		
		

		\begin{table}[h]
			\centering
			\renewcommand{\arraystretch}{1.25}
			\setlength\tabcolsep{4pt}
			\caption{Performance of the evaluated algorithms \\ on the mobile network dataset}
			\label{tab:mobile_perf}
			\begin{tabular}{l|c|c|c}
				& \multicolumn{2}{c|}{ACC}                     & NMI \\
				\cline{2-4}
				Alg.       & avg ($\pm$std)         & min - max           & avg ($\pm$std) \\
				\hline
				\ac{DEC}   & $0.7409$ ($\pm0.021$) & $0.7033$ - $0.7595$ & $0.8365$ ($\pm0.010$) \\
				\ac{ACAI}  & $0.7629$ ($\pm0.040$) & $0.7155$ - $0.8345$ & $0.8438$ ($\pm0.017$) \\
				\ac{IMSAT} & $0.4775$ ($\pm0.072$) & $0.3748$ - $0.5715$ & $0.5307$ ($\pm0.049$) \\
				\ac{DCCS}  & $0.8416$ ($\pm0.055$) & $0.7540$ - $0.9083$ & $0.8333$ ($\pm0.043$) \\
				\ac{DANCE} & $\mathbf{0.8923}$ ($\pm0.041$) & $\mathbf{0.8125}$ - $\mathbf{0.9305}$ & $\mathbf{0.8826}$ ($\pm0.035$) \\
			\end{tabular}
		\end{table}
	
		\ac{DEC} and \ac{ACAI}, the two generative algorithms show low average and peak performance, even in the lucky training cases with good \kmeans{} fits.
		We attribute this to the large amount of clustering-irrelevant information in the latent representation.
		The clusters formed by \kmeans{} and the \ac{DEC} mechanism ultimately incorporate this irrelevant information, which, in many instances, causes the encoded points to end up in the wrong cluster.
		On the other hand, the regularization through reconstruction seems to function well, as the deviation in accuracy for these algorithms is comparatively lower than the others.
		
		\ac{IMSAT} was not successful on the mobile network dataset, producing abysmal results.
		Originally, we have chosen \ac{IMSAT} to re-implement because contrary to being a discriminative algorithm, it did not utilize any domain-specific regularization methods such as image-transformations, rather, the regularization was done through the \ac{SAT} mechanism.
		\ac{SAT} disturbs the data on-the-fly during training, in a seemingly domain-agnostic manner, however, in order to calibrate the disturbance imposed by \ac{SAT}, \ac{IMSAT} uses a pre-calculated value $\epsilon$ for every datapoint.
		For the MNIST dataset, these $\epsilon$ values are the Euclidean distances between datapoints and their $10^{th}$ closest neighbors (calculated in the original data-space, each pixel is a separate dimension).
		The same calibration value calculated on the mobile network dataset does not seem to work well, for the reason being that the Euclidean distance is simply not that meaningful for our dataset as it is for MNIST, or images in general.
		The large variance in the position of important patterns in the sequences, caused by the arbitrary sequence framing, creates a large distance between even the same patterns shifted in time.
		We have tried to tune which neighbor to use for the $\epsilon$ calculation, but have seen no significant improvement.
		The bad performance could also be an indication of a mismatch in data complexity and the used neural net topologies, although we think the other algorithms are proof that the nets were at least capable of producing good results, if not optimal.
		
		\ac{DCCS} was the second most accurate algorithm on the mobile network dataset.
		\ac{DCCS} uses randomized data augmentation to separate 'categorical' features from 'style' features.
		These data augmentations are commonly used image transformations for the MNIST dataset: zooming, aspect ratio changes, brightness, hue and saturation changes.
		Zooming was quite straight-forward to implement for the mobile network dataset, and one could argue that such variation is probably present in the data: zooming in the temporal dimension is the equivalent of processes taking longer or shorter times, which manifests as the expansion or contraction of the generated patterns in the data.
		Aspect ratio changes do not apply to the mobile network dataset, as it behaves as an "image" which is a single pixel tall.
		We replaced value variations (brightness, hue and saturation changes) with a randomized offset and scaling on individual channels, tuning the parameters on the MNIST dataset to produce visibly similar images to the originally proposed image transformations.
		It seems that these augmentations were adequate for the mobile network dataset, as \ac{DCCS} proved to be quite accurate in it's clustering.
		We suspect that a fine-tuned net topology, as well as better tuned augmentations could further improve the performance of \ac{DCCS}, however, specifically this type of tuning is not possible if the user has only an unlabeled dataset available, the main premise of unsupervised learning.
		
\begin{figure}[h]
			\centering
			\subfloat[$Z_r$ ground truth]{
				\includegraphics[width=0.45\linewidth]{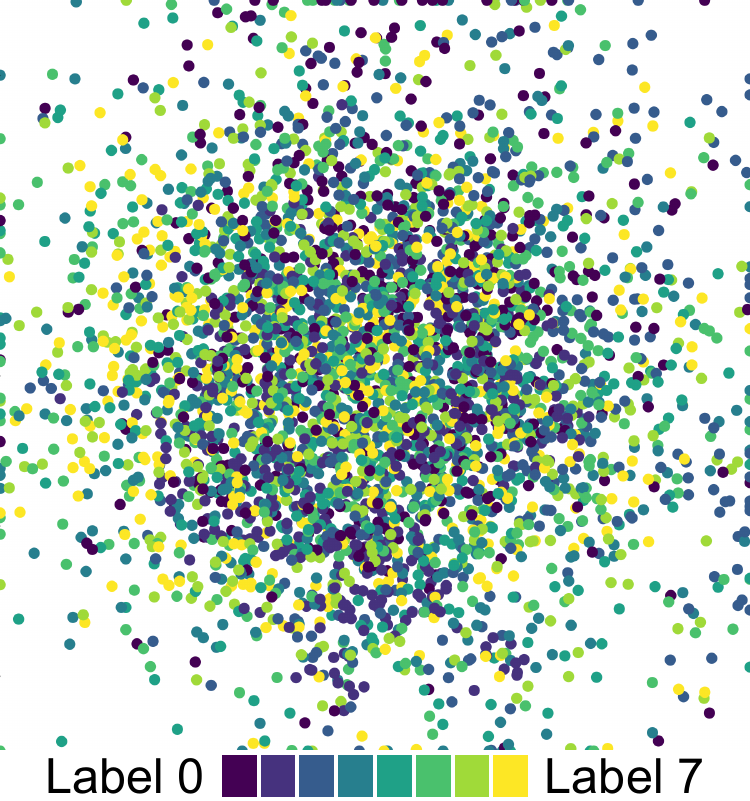}
				\label{fig:enc_mobile_zr_lab}
			}
			\subfloat[$Z_c$ ground truth]{
				\includegraphics[width=0.45\linewidth]{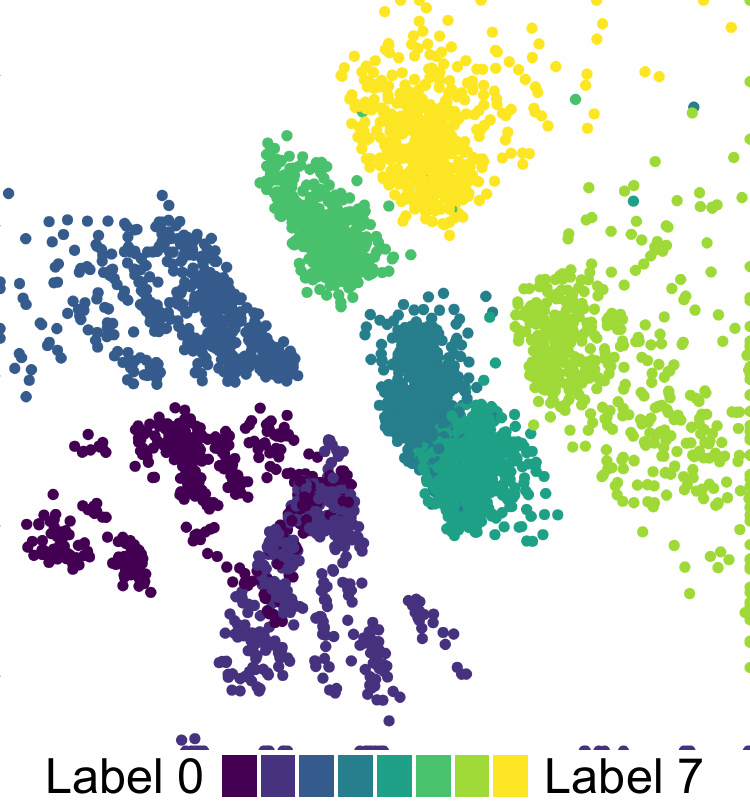}
				\label{fig:enc_mobile_zc_lab}
			}\\
			\subfloat[$Z_c$ density]{
				\includegraphics[width=0.45\linewidth]{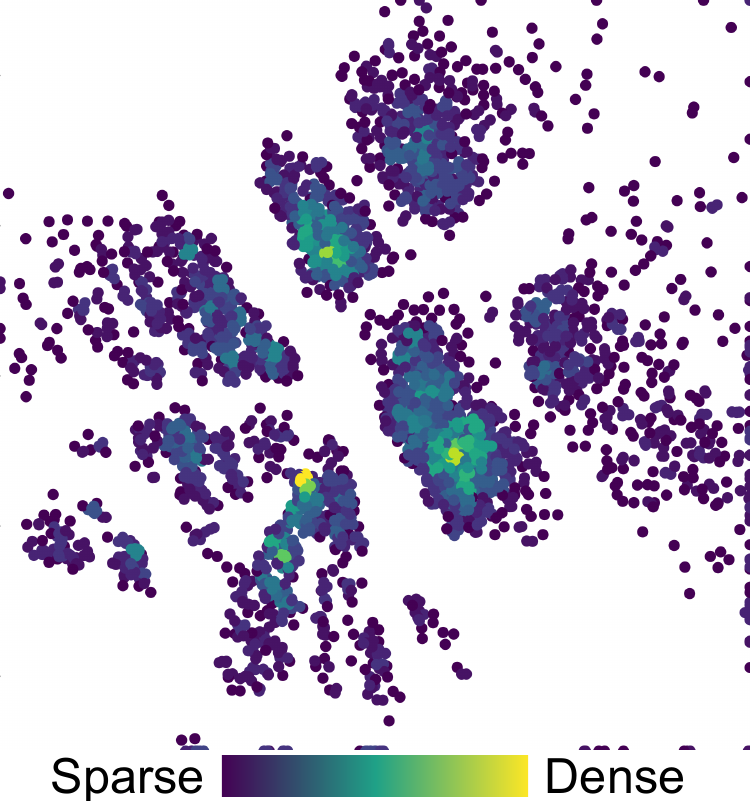}
				\label{fig:enc_mobile_zc_den}
			}
			\subfloat[\ac{RIM} initialization]{
				\includegraphics[width=0.45\linewidth]{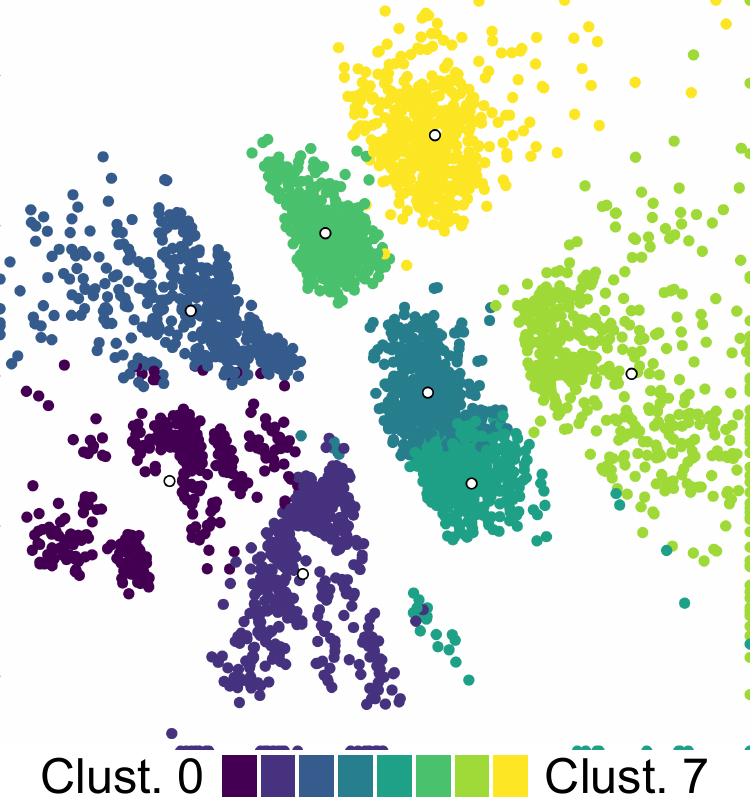}
				\label{fig:enc_mobile_rim_ass}
			}
			\caption{A typical \ac{DANCE} encoding of the mobile network dataset. The top two figures depict how well $Z_r$ follows the uncorrelated $p_g$ prior, as well as how well the ground truth classes separate in $Z_c$. The bottom two figures show the density of the encoded, points, and how well \ac{RIM} was able to find/determine the initial cluster centroids.}
			\label{fig:enc_mobile}
		\end{figure}

		\ac{DANCE} performed the best on the mobile network dataset, reaching the highest average, maximum, and most importantly the highest minimum accuracy.
		Our algorithm did not require hyper-parameter changes apart from changing $\beta_{cor}$; because of a higher reconstruction loss on the mobile network dataset, this balancing coefficient had to be increased in order for the decorrelator to have an effect on the encoding.
		This tuning can be done without any labeled data, solely by making sure that the decorrelator adversary converges close to $50\%$ accuracy (random guessing) when choosing between $Z$ and $Z'$ at the end of the training.
		A scatter-plot of the encoded datapoints and clustering can be seen on  Fig. \ref{fig:enc_mobile}.
		In order to further explore how this performance is achieved in \ac{DANCE}, we continue with an ablation study.
	
\section{Short Ablation Study}
	\label{sec:abl}

	It is important to see how much each of the \ac{DANCE} components contribute to the overall performance, which also helps in understanding the synergies between them.
	In the following ablation study, we examine every combination of the $3$ components evaluated on the mobile network dataset: the \ac{DAN}, the \ac{RIM} initialization and the \ac{DEC} cluster refinement.
	Without the \ac{DAN}, the autoencoder and the internal clustering steps worked in a single latent space, which was set to have the combined dimensionality of $Z_c$ and $Z_r$, resulting in $4$ dimensions.
	In the absence of \ac{RIM}, we utilized \kmeans{} to find the initial cluster centroids for \ac{DEC}.
	If \ac{DEC} was not used either, only \kmeans{} determined the final clustering.
	The results from the ablation study can be seen on Table \ref{tab:ablation_perf}.
	
	\begin{table}[h]
		\centering
		\renewcommand{\arraystretch}{1.25}
		\setlength\tabcolsep{4pt}
		\caption{The effect of the different dance components on performance \\ measured on the mobile dataset}
		\label{tab:ablation_perf}
		\begin{tabular}{c|c|c|c|c}
			\multicolumn{3}{c|}{}       & \multicolumn{2}{c}{ACC} \\
			\cline{4-5}
			\ac{DAN}        & \ac{RIM}        & \ac{DEC}        & avg ($\pm$std)                 & min - max  \\
			\hline
			                &                 &                 & $0.7135$ ($\pm0.0290$)         & $0.6518$ - $0.7515$ \\
           	                & $\surd$         &                 & $0.7295$ ($\pm0.0477$)         & $0.6375$ - $0.7775$ \\
           	\hlone{}        & \hlone{}        & \hlone{$\surd$} & \hlone{$0.7475$ ($\pm0.0391$)} & \hlone{$0.6720$ - $0.7985$} \\
           	                & $\surd$         & $\surd$         & $0.7598$ ($\pm0.0400$)         & $0.6845$ - $0.8010$ \\
           	\hline        
			$\surd$         &                 &                 & $0.7645$ ($\pm0.0665$)         & $0.7010$ - $0.9270$ \\
			$\surd$         & $\surd$         &                 & $0.8648$ ($\pm0.0396$)         & $0.7850$ - $0.9120$ \\
			$\surd$         &                 & $\surd$         & $0.7823$ ($\pm0.0615$)         & $0.7145$ - $0.9255$ \\
			\hltwo{$\surd$} & \hltwo{$\surd$} & \hltwo{$\surd$} & \hltwo{$0.8923$ ($\pm0.0410$)} & \hltwo{$0.8125$ - $0.9305$} \\
		\end{tabular}
	\end{table}

	Without any of the $3$ components, the algorithm is simply \kmeans{} run on an autoencoder-formed latent encoding.
	This setup is often used as a baseline for deep clustering, with the premise that the algorithm should greatly improve upon these results.
	Using \ac{RIM} instead of the \kmeans{} clustering does not bring tangible benefits, probably because the irrelevant information in the latent encoding hides the otherwise sparsely populated cluster boundaries \ac{RIM} is looking for.
	Using \ac{DEC} with a \kmeans{} initialization is basically the originally proposed \ac{DEC} algorithm (highlighted with blue), however, the results are a little worse than what we have shown in Table \ref{tab:mobile_perf}, because in this case \ac{DEC} is operating with a lower dimensional latent space.
	Adding \ac{RIM} as an initialization for \ac{DEC} once again does not improve performance meaningfully, for the same reason \ac{RIM} was not greatly beneficial by itself.
	
	Using \ac{DAN} and clustering with \kmeans{} only in $Z_c$ already improves the average accuracy as much as the other two components combined, but more importantly improves peak accuracy by a great margin.
	This is because the decorrelated encoding in $Z_c$ maps clusters in a compact manner, without many datapoints mixed into wrong clusters.
	\kmeans{}, though unreliably, sometimes fits these clusters well, resulting in high peak accuracy.
	Using \ac{RIM} instead of \kmeans{} for clustering greatly improves minimum and average accuracy, because in $Z_c$, without most of the clustering-irrelevant information, the sparse cluster boundaries stand out, and \ac{RIM} is able to find these reliably.
	Not using \ac{RIM} but using \ac{DEC} once again loses these benefits, only retaining the high peak accuracy achieved through the \ac{DAN} decorrelation.
	Finally, with all components combined, we arrive at the complete \ac{DANCE} algorithm (highlighted with green), where \ac{DEC} is able to exert its full benefits on the clustering, improving worse and average accuracy by quite a significant margin without losing its capability to maximize peak accuracy.

\section{Conclusion}
	
	In this paper, we discussed state-of-the-art deep clustering algorithms, splitting them into two groups: generative and discriminative methods.
	Although discriminative methods seem to be the peak performers in image clustering, their highly tuned nature and assumptions about the data make them hard to apply to mobile network data.
	Reasoning that generative algorithms seem to be more domain-agnostic, we have proposed our own generative deep clustering algorithm, \ac{DANCE}, with the core idea of isolating clustering-relevant features in the latent space.
	We have evaluated \ac{DANCE} and other state-of-the-art algorithms' performance on an image- and a mobile network dataset, while also providing an ablation study to highlight the significance of the different components of our algorithm.
	\ac{DANCE} achieved good performance on the image dataset, and excellent performance on the mobile network dataset, surpassing its competitors by a sizable margin.

	In real-world applications, clustering algorithms require a great deal of expertise to use. 
	In the hands of a less experienced user, or somebody who does not have the resources, time, or a labeled dataset to fine-tune these algorithms for the specific use-case, simplicity, usability and reliability play a far bigger role than peak performance in the overall usefulness of the algorithm.
	
	As a closing remark, we would like to highlight the shared idea between \ac{DCCS} and our \ac{DANCE} algorithm; the concept of separating latent features into clustering-relevant and clustering-irrelevant sets.
	Although both implementations show various advantages and disadvantages in different data domains, at the least we can say that the concept itself is very promising, and could be an interesting topic for future research.





\bibliographystyle{IEEEtran}
\bibliography{dan_tnnls_bibliography}{}

%

%
%

\end{document}